# GRAPH NEURAL NETWORKS FOR TRAFFIC FORECASTING

**João Rico*[1,2,3], José Bareteiro[1,2], Arlindo Oliveira[2,3]**

[1]Laboratório Nacional de Engenharia Civil, Av. do Brasil 101, Lisbon, Portugal
[2]INESC-ID, R. Alves Redol 9, Lisbon, Portugal
[3]Instituto Superior Técnico, University of Lisbon, Lisbon, Portugal
jmrico@lnec.pt, jbareteiro@lnec.pt, arlindo.oliveira@tecnico.ulisboa.pt

## 1 Introduction

The world's urban population will increase from 4.2 billion to 6.7 billion by 2050 as estimated by the United Nations [1]. In spite of the accompanying social evolution and benefits, the rapid rate of urbanization has significant social, economic and environmental costs associated, including air and water pollution, unsustainable energy consumption, toxic waste disposal, inadequate urban planning, decreased public health and safety, social vulnerability and community disruption. Notably, the mobility of passengers and freights in most large cities of the world is not yet sustainable. Urbanization has given rise to traffic congestion, increase of transports needs, ineffective accessibility, and reduced productivity. In particular, traffic congestion costs billions of dollars per year due to lost time, air pollution, and wasted fuel. In 2017, in the United States alone, traffic congestion induced a total of 8.8 billion hours of travel delay and 12.5 billion liters of extra fuel consumption, corresponding to a congestion cost of 166 billion dollars [2]. Efforts to develop solutions to the challenges brought by traffic congestion have focused on three avenues: championing transport alternatives, enlarging the infrastructure, and managing traffic flows [3]. While championing transport alternatives is mostly a public policy issue, and geographical and social constraints limit the increase of the size of infrastructure, the potential to efficiently manage traffic flows has increasingly become one important solution to traffic congestion. Today, the exponential increase of available data in cities and the growth of computing capabilities represents a crucial opportunity to tackle these challenges by leveraging innovative and integrated solutions. These include the development of intelligent transportation systems (ITS), smart vehicle sharing systems and home automation, and smart grids and energy solutions - all of which fall under the umbrella of the recent area of urban computing [4].

A core component of ITS is traffic forecasting. Its goal is to measure, model and predict traffic conditions in real-time, accurately and reliably, in order to optimize the flow and mitigate the congestion of traffic, and to respond adequately to other problems such as traffic light control, time of arrival estimates, and planning of new road segments. However, this is a very challenging problem due to several important factors. The successful forecasting of traffic conditions requires adequately handling heterogeneous data (e.g., integrating loop counter and floating car data) with complex spatio-temporal dependencies which are typically sparse, incomplete and high-dimensional. In addition, it requires computing in real-time and the inclusion of external factors such as weather conditions and road accidents.

This review is focused on the recent developments and applications of graph neural networks, a new family of deep learning models, to road traffic forecasting. In Section 2, we present a succinct history of traffic forecasting including traditional approaches and deep learning models that have had a great deal of success. Section 3 presents and discusses graph neural networks (GNN) and Section 4 reviews the literature of traffic forecasting based on GNN. We conclude, in Section 5, with a discussion of open challenges and research opportunities.

## 2 Traffic forecasting

Traffic forecasting, simultaneously a subproblem of spatiotemporal data mining and urban computing, aims to predict traffic conditions at a future time in a network given a sequence of historical traffic states. Typical variables predicted include speed, volume, density, flow, congestion condition, and occupancy. Most works in this review follow an equivalent or similar problem formulation [5]–[8]. In this formulation, a graph $G = (V, E, W)$ represents some underlying structure of the problem (such as the spatial distances between sensors or the similarities between their time series) that we are trying to leverage as an inductive bias of the model, and where V is the node set, E the edge set, and W the weighted adjacency matrix. Denoting $X_t{}^i$ as the values of all the features of node $i$ at time $t$, we aim to learn a function $F$ that given a sequence of $M$ historical time steps predicts the next $N$ time steps:

$$F([X_{t-M+1}, \dots, X_t]; G) = [X_{t+1}, \dots, X_{t+N}]. \quad (1)$$

## 2.1 Traffic data

Urban data can be classified according to their structures and spatiotemporal properties [9], as Figure 1 illustrates. We can distinguish datasets with respect to their spatiotemporal properties: spatiotemporal static data, spatial static but temporal dynamic data, or spatiotemporal dynamic data. They can also be classified with respect to their data structures in one of two types - point-based and network-based. In addition, urban data can also be classified according to its sources, such as geographical data, traffic data, and social network data, to name a few, and each of these can be further subclassified. Focusing on traffic data, this source can include loop detector data, floating car data, mobile phone data, call detail records, surveillance cameras, bike-sharing data, taxi records, mobile phone location data, parking records, and mobile apps' logs.

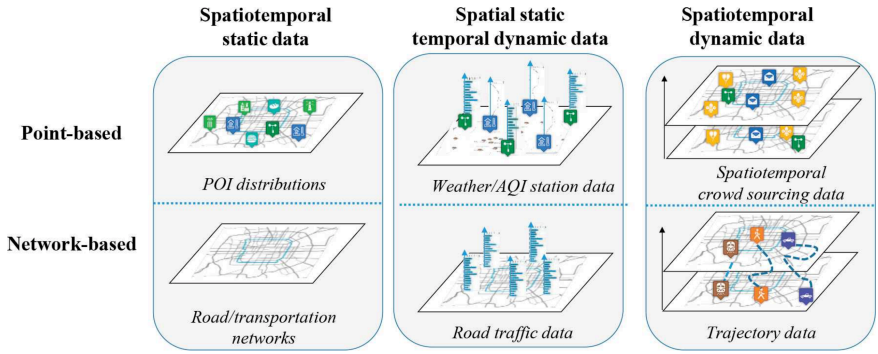

Figure 1: Six types of urban data [9]. Reprinted with permission.

## 2.2 Traditional approaches to traffic forecasting

In the last decades, traffic forecasting methods have been developed by researchers from different fields such as transportation systems [10], economics [11], statistics, and machine learning [12]. The methods can be divided into two categories: the knowledge- or model-driven approach and the data-driven approach. Knowledge-driven approaches usually aim at modeling and explaining the transportation network through differential equations and numerical simulation [13], [14]. Although the models can reproduce real traffic conditions fairly accurately, they require prior knowledge and detailed modeling, are not easily transferable to other cases, and require significant computational resources.

Traditional data-driven methods used in traffic forecasting can be divided into two categories: methods that do not model the spatial dependency and models that do model this dependency. The former category includes the Historical Average (HA) model, the auto-regressive integrated moving average (ARIMA) [15] and seasonal ARIMA (SARIMA) [16], K-nearest Neighbour (KNN) [17], support vector regression [18], hidden Markov models [19], among others. These approaches often require careful feature engineering and also rely on data to satisfy certain assumptions, such as stationarity. However, real data is often too complex and violates often these assumptions, leading in many cases to poor performance. Approaches that model the spatial dependency include Vector ARIMA, Spatiotemporal ARIMA and Spatiotemporal HMM [20]–[23]. Again, these methods perform poorly since they are not complex enough to model the non-linearity and non-stationarity of the data.

## 2.3 Deep learning approaches to traffic forecasting

Deep learning methods [24] are a class of machine learning methods that learn multiple layers of representations by composing increasingly more complex non-linear features on the upper layers by combining simpler features from the lower layers. This learning of complex representations is done mostly automatically and does not depend on a human doing manual feature engineering, which would require time and expert domain knowledge. In recent years, deep learning represents the state of the art in fields such as image recognition [25], natural language understanding [26], drug discovery [27], recommendation systems [28], and board and video games playing [29]. This success is mainly due to their representation power (as explained above) and to the fact that there exists an efficient method for training them (namely gradient descent through error backpropagation).

As in other domains, the application of deep learning methods to traffic forecasting has been very successful and has produced state-of-the-art results. Some of the earlier architectures did not model the spatial dependency, and used standard feedforward networks or deep belief networks [30] as well as recurrent neural networks such as long short-term memory (LSTM) and the gated recurrent unit (GRU) [31]–[33]. However, these models still fail to model the complex spatial dependencies that exist in traffic problems. Several models were proposed that aim at capturing these dependencies, using convolutional neural networks (CNN) [34] or a combination of CNN and LSTM [35], [36]. Still, since CNNs are mainly suited for data embedded in grid-line Euclidean spaces, they are not the natural architecture to real-world road networks. The next sections expand on how models using a graph neural network improved on these architectures.

## 3 Graph neural networks

The convolutional operator used in CNNs is very powerful but is limited to standard grid data, i.e., to data that originates in regular, two-dimensional, tri-dimensional, or higher-dimensional Euclidean spaces. Since various important machine learning problems involve tasks on graph structured data such as node classification [37] or molecule generation [38], researchers have developed a family of deep learning models that can leverage this inductive bias [37], [39]–[41]. These models are usually called graph neural networks (GNN) because they are neural networks that naturally handle graph data. In this section, we briefly review the area of graph neural networks and some of the most relevant works, and refer the reader to other reviews for more details and specific perspectives on this topic [41]–[46]. In Section 3.1, we categorize and describe several GNN models, and in Section 3.2 we list current open-source libraries for implementing GNNs.

### 3.1 Models

In this section, we describe several of the most relevant GNNs models developed so far. As shown in Table 1, we categorize GNNs into four main types, namely recurrent graph neural networks, convolutional graph neural networks, graph autoencoders and graph attention networks.

| Category | References |
|---|---|
| Recurrent Graph Neural Networks | [39], [47]–[50] |
| Convolutional Graph Neural Networks | [37], [40], [51]–[53] |
| Graph Attention Networks | [54], [55] |
| Graph Autoencoders | [56]–[61] |

Table 1. Categorization of graph neural network models and representative publications.

*3.1.1 Recurrent graph neural networks*

Extending previous work [62], [63], the Graph Neural Network model (GraphNN) [39] was the first neural network model that could process general types of graphs (eg, directed, undirected, cyclic, or acyclic). The fundamental concept of the GraphNN model is that every node $v$ can be represented by a low-dimensional state vector $h_v$ and be defined by its features and by its neighbours based on an information diffusion mechanism from every node to its neighbors. The goal is to learn this representation which can be fed to an output function $g$, called the local output function, resulting in an output value or label $o_v$, for regression or classification, respectively. The model also defines the local transition function $f$, a parametric function, to be learned alongside $g$ - both shared by all nodes. Together, these node representation and output are defined as:

$$h_v = f(x_v, x_{co[v]}, h_{ne[v]}, x_{ne[v]}),$$ (1)

$$o_v = g(h_v, x_v),$$ (2)

where $x_v, x_{co[v]}, h_{ne[v]}, x_{ne[v]}$ are the features of $v$, the features of its edges, and the states and features of its neighborhood, respectively.

Denoting the vectors constructed by stacking all the states, all the features, all the node features and all the outputs by $H, X, X_N, O$ equations (1) and (2) can be rewritten compactly as

$$H = F(H, X),$$ (3)

$$O = G(H, X_N),$$ (4)

where $F$ and $G$ are the stacked versions of $f$ and $g$ for all nodes, respectively.

If $F$ is a contraction map, Banach's Fixed Point Theorem [64] guarantees the existence and uniqueness of the system of equations (3) and (4). This suggests iteratively updating the following equation:

$$H^t = F(H^{t-1}, X)$$ (5)

where $H^t$ is the $t$-th iteration of $H$, and $H^0$ can be initialized randomly.

Figure 2 shows the intuition behind equations (1) and (5): to learn a node representation by propagating information from its neighbors, iteratively, until convergence.

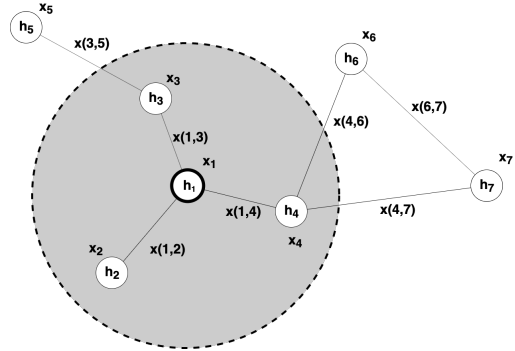

$$h_1 = (x_1, x_{(1,2)}, x_{(1,3)}, x_{(1,4)}, h_2, h_3, h_4, x_2, x_3, x_4)$$

Figure 2: Example graph of how GraphNN propagates information from the states and features of the immediate neighborhood of a node.

To learn the functions $f$ and $g$, GraphNN defines a loss function at the supervised nodes, and uses a gradient descent scheme: iteratively updating $H^t$ until a convergence criterion is reached; calculating the gradients of the loss with respect to the weights of $f$ and $g$; and updating the weights.

Although it is a simple and powerful model, GraphNN has several limitations. Most importantly, it is computationally inefficient to compute the iterations for the fixed point and requiring $F$ to be a contraction map limits the modeling capacity of the approach, including the long range dependencies of nodes (see appendix A of [47]).

The Gated Graph Neural Network (GatedGNN) [47] is another recurrent graph neural network, and it improves on some of the drawbacks of GraphNN. This model modifies the original GraphNN substituting the recurrence function in equation (1) for a Gated Recurrent Unit (GRU) [65]. GatedGNN uses backpropagation through time (BPTT), unrolling the recurrence for a fixed number of T steps. While for a large graph this can present a drawback by requiring a large amount of memory to store the intermediate states computed over all nodes, on the other hand this removes the requirement for a contraction map in order to guarantee convergence. The model recurrence is defined as follows

$$a_v{}^t = A_{v:}[h_1{}^{t-1} \dots h_N{}^{t-1}] + b, \qquad g_v{}^t = tanh(Wa_v{}^t + U(h_v{}^{t-1} \odot h_v{}^{t-1}),$$

$$z_v{}^t = \sigma(W^z a_v{}^t + U^z h_v{}^{t-1}), \qquad h_v{}^t = (1 - z_v{}^t) \odot h_v{}^{t-1} + (z_v{}^t \odot g_v{}^t), \qquad (6)$$

$$r_v{}^t = \sigma(W^r a_v{}^t + U^r h_v{}^{t-1}),$$

where $A$ is the modified adjacency matrix composed of two submatrices, one for outgoing edges, the other for incoming edges. Figure 3 shows the parameter tying and sparsity structure in the adjacency matrix, as well as an example of the unrolling of the recurrence for one time step. $A_{v:}$ corresponds to the two columns of the submatrices $A^{(out)}$ and $A^{(out)}$ referring to node $v$, and $z$ and $r$ correspond to the update and reset gates of the GRU.

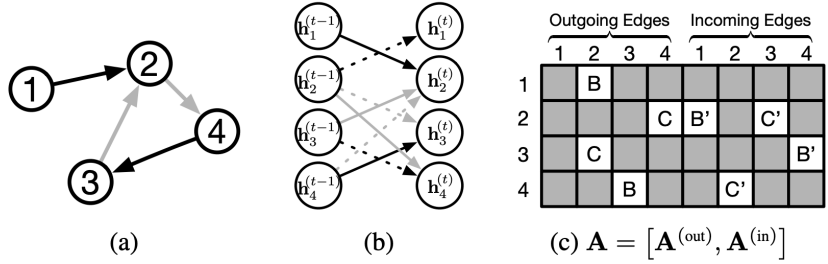

(a)             (b)             (c) $\mathbf{A} = \left[\mathbf{A}^{(out)}, \mathbf{A}^{(in)}\right]$

Figure 3: (a) Example graph, where edge color denotes edge type. (b) Unrolling of one time step. (c) Parameter tying and sparsity in adjacency matrix. Letters denote edge types. B′ corresponds to the reverse edge of type B. B and B ′ denote distinct parameters [47]. Reprinted with permission.

Other recurrent graph neural networks have been developed, including the Graph Echo State Network [66], Graphrnn [50] and Stochastic Steady-state Embedding [49].

### 3.1.2 Convolutional graph neural networks

The generalization of the very successful CNN to non-Euclidean data such as graphs and manifolds has followed two distinct avenues, the so-called spectral methods and spatial methods. Briefly, spectral based methods define the convolution based on the graph Laplacian and use the corresponding Fourier basis in the spectral domain, and spatial methods use a localized parameter-sharing filter on the neighbors of each node, in essence, aggregating local features.

Spectral Convolutional Neural Networks (SpectralCNN) [51] was the first model to introduce the convolution for graph data. Given an undirected graph its normalized graph Laplacian is defined as

$$L = I_n - D^{-1/2} A D^{-1/2}, \qquad (7)$$

where $A$ is the adjacency matrix and $D$ is the diagonal matrix of node degrees. The graph Laplacian is a real symmetric positive semidefinite matrix which implies that it can be factorized as

$$L = U \Lambda U^T, \qquad (8)$$

where $\Lambda$ is the diagonal matrix of eigenvalues - the spectrum - and $U$ is the matrix of eigenvectors. Given a graph signal $x \in R^n$ where $x_i$ is the feature of the $i$-th node. The graph Fourier transform $\mathcal{F}$ - and its inverse $\mathcal{F}^{-1}$ are defined by

$$\mathcal{F}(x) = U^T x = \hat{x}, \qquad (9)$$

$$\mathcal{F}^{-1}(\hat{x}) = U^T \hat{x}. \qquad (10)$$

The graph convolution operation of a graph signal x with a filter g on a graph G is defined as

$$x * g = \mathcal{F}^{-1}(x)(\mathcal{F}(x) \odot \mathcal{F}(x)),$$

$$= U(U^T x \odot U^T g) \qquad (11)$$

where $\odot$ is the Hadamard product. The equation above can be simplified, by considering a filter $g = diag(\theta)$ parametrized by $\theta \in R^n$ in the Fourier domain, to the following expression

$$x * g = U g U^T x, \tag{12}$$

where filter $g$ can be understood as a function of eigenvalues of the Laplacian, that is $g(\Lambda)$.

This model has several drawbacks. It is computacional inefficient since multiplication by $U$ is $O(n^2)$ and calculating the eigen-decomposition is $O(n^3)$. In addition, the dimensionality of the trainable filter depends on the number of nodes of the graph, and the filters are non-localized depending on the graph structure - if a graph changes at all, so will its eigenvalues and eigenvectors.

Chebyshev Spectral Convolutional Neural Networks (ChebNet) [40] proposes to approximate the filter by Chebyshev polynomials of the diagonal matrix of eigenvalues, that is,

$$x * g = U(\sum_{i=0}^{K-1} \theta_i T_k(\Lambda^*)) U^T x,$$

$$= \sum_{i=0}^{K-1} \theta_i T_k(L^*) x, \tag{13}$$

where $L^* = 2L/\lambda_{max} - I_N$ and $\lambda_{max}$ is the maximum eigenvalue. The Chebyshev polynomials are defined by the recursion $T_k(x) = 2x T_{k-1}(x) - T_{k-2}(x)$ with $T_0(x) = 1$ and $T_1(x) = x$. In this setting, the computational complexity reduces to $O(KM)$ where $K$ is the polynomial order of the expansion above and $M$ is the number of edges, rendering the model linear with respect to the graph size. This also renders the filters spatially localized, since for an expansion of degree $K$ each node only receives information from a node at maximum $K$ hops away.

Graph Convolutional Networks (GCN) [37] extends ChebNet by limiting the Chebyshev expansion to only the first order, $K = 1$, and employing some additional modifications. It assumes $\lambda_{max} = 2$, simplifying equation (13) to

$$x * g = \theta_0 x - \theta_1 D^{-1/2} A D^{-1/2} x. \tag{14}$$

Setting $\theta_0 = -\theta_1$ and substituting $A$ for $A^* = A + I_N$, equation (12) can be written as

$$x * g = \theta(I_N + D^{*-1/2} A^* D^{*-1/2} x. \tag{15}$$

Generalizing to the signal to $C$ input channels and $F$ filters for feature maps, the graph convolution can be written as

$$Z = D^{*-1/2} A^* D^{*-1/2} \theta, \tag{16}$$

where $Z \in R^{N \times F}$, and $\Theta \in R^{C \times F}$ is a matrix of filter parameters.

ChebNet and GCN bridge the gap between spectral- and spatial based methods. Equation (16) can be expressed as

$$h_v = f((\sum_{u \in \{N(v) \cup v\}} A_{v,u} x_u)\Theta), \tag{16}$$

which can be interpreted as a spatial-based method, each node aggregating information from its neighbors.
Neural Fingerprints (Neural FPs) [53] generalized GNNs by using different learning parameters for nodes with different degrees, as follows:

$$x = h_v^{t-1} + \sum_{i=1}^{|N(v)|} h_i^{t-1} \theta_i T_k(\Lambda^*)) U^T x, \tag{17}$$

$$h_v^t = \sigma(x W_t^{|N(v)|}), \tag{18}$$

where $W_t^{|N(v)|}$ is the weight matrix for nodes of degree $N$ at layer $t$.

Another interesting model is the dual graph convolutional network (DGCN) [67] which jointly uses two convolutions: one based on the adjacency matrix is the same as equation (16), and the second one is based on a diffusion process to capture global relationships by substituting the adjacency matrix $A$ with the positive pointwise mutual information (PPMI) matrix [52].

### 3.1.3 Graph Attention Networks

Attention mechanisms [68], [69] are a recent development in deep learning that has had success in tasks such as machine translation. Figure 4 illustrates the intuition of why attention mechanisms could be helpful in modelling the neighborhood of a node, namely by leveraging the type of each neighbor to assign different importance weights.

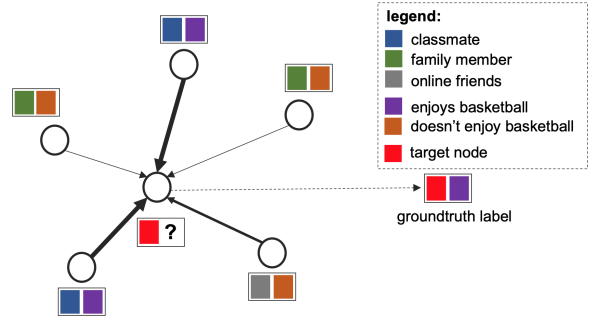

Figure 4: The type of each neighbor is used to assign attention, and the link size denotes how much attention to apply to each neighbor. This example illustrates how concentrating our attention on the node's classmates, would improve our prediction of the enjoyment of the activity [70]. Reprinted with permission.

Graph Attention Networks (GAT) [54] was the first model to incorporate an attention mechanism in GNNs by removing the requirement that all the neighbors of a node contribute with equal or pre-defined weights. Rather, the relative weights between two connected weights are learned as follows:

$$h_v^t = \sigma(\sum_{u \in \{N(v) \cup v\}} \alpha_{vu} W^{t-1} h_u^{t-1}), \qquad (19)$$

where $\alpha_{vu}$ is the attention coefficient of node $j$ to $i$ defined as

$$\alpha_{vu} = softmax(g(a^T[W^{t-1}h_v||W^{t-1}h_u])), \qquad (20)$$

where $a$ is a vector of learnable parameters and $g$ is a LeakyReLU unit.

GAT also proposes incorporating a multi-head attention mechanism [69], and have the important advantage of being efficient since the computation of the node-neighbor pairs is parallelizable and because it can naturally be applied to nodes with a different number of neighbors.

Gated Attention Network (GAAN) [55] is an extension of GAT which, in addition to the multi-head attention mechanism, also incorporates a self-attention mechanism which replaces the averaging operation of GAT with an attention score on each attention head. For more details and different taxonomies of attention models in graphs, we refer the reader to a comprehensive review [70].

### 3.1.4 Graph Autoencoders

Graph neural networks have also been used in unsupervised learning settings. In particular, successful deep learning models such as autoencoders [71] and variational autoencoders [72] have been generalized to handle graph data. These models are usually developed to either learn network embeddings, such as Structural Deep Network Embedding [56] and Variational Graph Autoencoder [57], or to generate new graphs such Deep Generative Model of Graphs [58] and Graph Variational Autoencoder [59].

Some models have also incorporated adversarial training, in particular Generative Adversarial Networks (GAN) [73]. In a GAN, two networks compete against each other: the generator network is optimized to produce realistic samples of data that the discriminator network tries to distinguish from real data. GANs have been used together with GCN for generation of molecular graphs [60], link prediction, node clustering, and graph visualization [61].

### 3.1.5 Frameworks

In addition to the different types of GNN models that have been developed so far, some of which we describe above, there has been an effort to unify GNNs under a general and common framework. Non-local Neural Networks (NLNN) [74] unifies different self-attention mechanisms, and Message Passing Neural Networks (MPNN) [75] generalizes several GNN approaches using a message passing scheme. In MPNNs, messages $m_v{}^t$ at timestep $t$ are passed from each node $v$ as follows:

$$m_t{}^{t+1} = \sum_{u \in N(v)} M^t(h_v{}^t, h_v{}^t, e_{vu}), \tag{21}$$

$$h_t{}^{t+1} = U^t(h_v{}^t, m_v{}^{t+1}), \tag{22}$$

where $U^t$ is the vertex update function, $M^t$ is the message function and $e_{vu}$ represents the features of the edge from node $v$ to $u$.

The Graph Networks (GN) model [41] generalizes both the NLNN and MPNN frameworks and other types of GNNs. It defines a GN block containing three update functions and three aggregation functions as follows:

$$\begin{aligned}
e_k{}' &= \varphi^e(e_k, h_{rk}, h_{sk}, u), & e_i{}^{*\prime} &= \rho^{e \to h}(E_i'), \\
h_i{}' &= \varphi^h(e_i{}^{*\prime}, h_i, u), & e^{*\prime} &= \rho^{e \to u}(E'), \\
u' &= \varphi^u(e^{*\prime}, h^{*\prime}, u), & h^{*\prime} &= \rho^{h \to u}(H'),
\end{aligned} \tag{23}$$

where $\rho()$ are the message passing functions and $\varphi()$ the update functions.

## 3.2 Software

The research and application of GNNs has been greatly enhanced by the development and publishing of several open-source libraries, specifically written to handle GNNs. These are built on top of already powerful deep learning frameworks, such as PyTorch [76] and Tensorflow [77]. These GNN libraries can greatly reduce the time of testing and deployment of new models, by providing useful abstractions that simplify the code required while automatically handling several low-level optimizations, such as scaling, parallelization and taking advantage of sparse structure. It is also common for authors to release implementations of their models in these libraries, which are, in turn, often integrated in these libraries in a ready-to-use manner.

| Library | Reference | DL Framework | URL |
|---|---|---|---|
| DGL | [78] | MXNet, Pytorch | https://github.com/dmlc/dgl |
| Euler | [79] | Tensorflow | https://github.com/alibaba/euler |
| Graph Nets | [41] | Tensorflow | https://github.com/deepmind/graph_nets |
| PyTorch Geometric | [80] | Pytorch | https://github.com/rusty1s/pytorch_geometric |

Table 2: Graph deep learning libraries.

## 4  Graph neural networks for traffic forecasting

The connection between graphs and roads or maps is as old as graph theory, dating back to Euler's formulation and solution to the Königsberg problem. Nodes can represent road intersections and edges can model the road segments between them. Or, almost as naturally, nodes can represent points on a road segment where a traffic variable is being measured by a sensor and edges represent some relationship between these locations, such as the shortest path distance between them. The addition of a temporal component can be just as natural, as Figure 6 illustrates. Typically, each node in a graph would define a time-series of a feature vector, such as the history of traffic speed and traffic flow at a certain location.

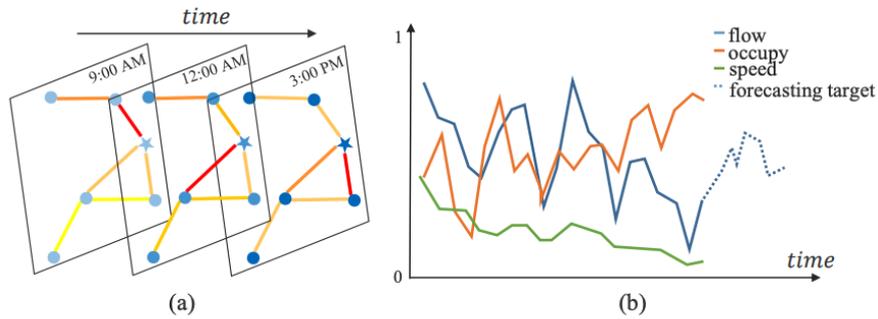

Figure 6: (a) Typical spatio-temporal structure of traffic data with each time slice forming a graph. (b) Representation of a three-dimensional time series at a certain node and a forecast of one of these features. [8]. Reprinted with permission.

Traffic forecasting based on graph neural networks can be viewed as the problem of extending the models of the previous section and the categories of Table 1 to the temporal domain. In principle, combining the many types of GNNs approaches with the immense diversity of existing time series analysis models and forecasting methods offers a wide space of possibilities.

In this section, we will present and compare the graph neural network models for traffic forecasting that have been developed so far. These have in common a graph neural network as a centerpiece of the model and they vary along most modern deep learning paradigms, combining recent techniques that have been found powerful in other domains of applications. These include attention mechanisms [8], [55], multi-graphs [81], dynamic graphs [82],  diffusion kernels [83], graph wavenet [84], inception models [85],  deep graph infomax [86], residual nets [87], wavelet transforms [88], among others [6], [7], [89]–[95]. In section 4.1. we review these models and compare different approaches, and in Section 2 we describe typical datasets in which these models are evaluated.

### 4.1 Models

Table 3 lists the GNNs models developed for traffic forecasting so far, which we describe below. We start by describing Spatio-Temporal Graph Convolutional Networks (STGCN) [92] and Diffusion Convolutional Recurrent Neural Networks (DCRNN) [83]. As of the writing of this review, these two models stand out as the most cited and most often compared to when developing a new model, in part because they are some of the models first developed while also representing opposite approaches in some aspects.

STGCNs combine the spectral-based ChebNets [40] with 1D-CNNs to model and predict traffic speed at various locations where sensors are located. The GCN and the CNN operate alternately along several layers, with the GCN capturing the spatial dependency while the CNN captures the temporal dependency.  Figure 7 shows an illustration of scheme similar to the one used by STGCN, in which a CNN is combined with a GCN.

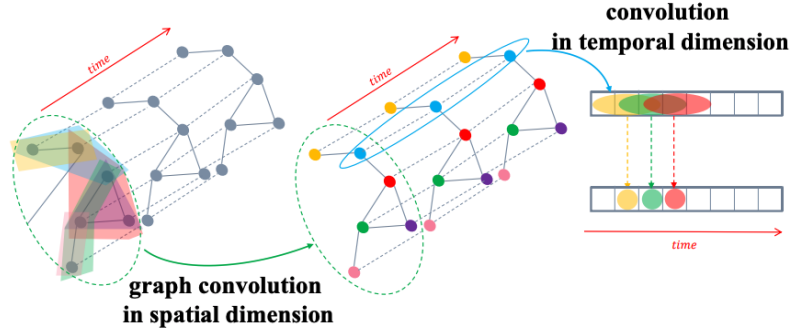

Figure 7: Illustration of how to combine GNN and CNN to capture the spatial and temporal dependencies, respectively [8]. Reprinted with permission.

DCRNNs take the opposite direction: they use RNNs instead of 1D-CNN to model the time dependencies and, in addition, use a spatial-based GCN for the spatial dependency instead of a spectral-based one. This is done by incorporating the GCN inside a GRU. Based on Diffusion-Convolution Neural Networks [96], DCRNN uses the probability transition matrix $P = D^{-1}A$ to define the graph convolution as follows:

$$H = \sum_{k=0}^{K} f(P^k X W^{(k)}), \tag{24}$$

where $f()$ is an activation function and $W^{(k)} \in R^{D \times F}$. Effectively, DCRNN assumes that by passing information from node to node the network can reach an equilibrium state which is captured by the probability transition matrix. DCRNN also uses an enconder-decoder architecture to predict future timesteps.

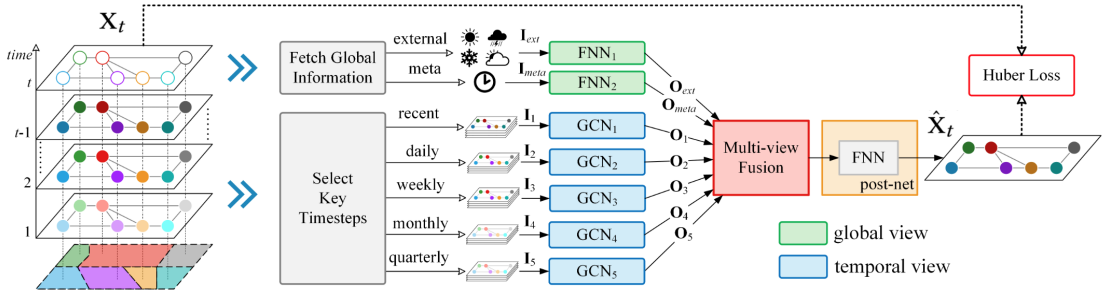

Figure 8: Architecture of MVGCN, displaying the views capturing different timescale dependencies, and their fusion in the final layer. [90]

A different and promising approach has been proposed by several recent models, including Multi-View Graph Convolutional Networks (MVGCN) [90], Multi Residual Recurrent Graph Neural Networks (MRes-RGNN) [87], Motif-based Graph Convolutional Recurrent Neural Network (Motif-GCRNN) [88] and Spatial-Temporal Graph Inception Residual Networks (STGI-ResNet) [85]. In these approaches, typical repeating time patterns are grouped in order to leverage the inductive bias we know is present in typical traffic, such as daily or weekly periods. As an illustration, if one is predicting traffic speed at 9am on a friday, in principle, one can make a better prediction by considering not only the last previous hours, but also, for example, the traffic at 9am in the previous four days of the week, and in the previous fridays, at 9am. These different views are then fused in the last layers of the architecture. This fusion layer is also learnable. Figure 8 shows the architecture of MVGCN which exemplifies this approach. As shown in the figure, these approaches can also typically handle exogenous factors such as weather and irregular events (e.g., road accidents), that have a significant impact in urban traffic.

| Model | Ref. | Scope | Predicts | Data source | Datasets | Open dataset? | Code available? |
|---|---|---|---|---|---|---|---|
| ST-GCN | [92] | Fw, Ur | S | L | BJER4, PeMS | ✗, ✓ | ✓ |
| DCRNN | [83] | Fw | S | L | METR-LA, PeMS | ✓ | ✓ |
| MRes-RGNN | [87] | Fw | S | L | METR-LA, PeMS | ✓ | ✗ |
| TGC-LSTM | [7] | Fw, Ur | S | L, FCD | LOOP, INRIX | ✓, ✗ | ✗ |
| ASTGCN | [8] | Fw | F, S | L | PeMSD4, PeMSD8 | ✓ | ✓ |
| STDGI | [86] | Fw | S | L | METR-LA | ✓ | ✓ |
| MVGCN | [90] | Ur | F | FCD | TaxiNYC, TaxiBJ, BikeDC, BikeNYC | ✓ | ✗ |
| DST-GCNN | [82] | Fw, Ur | S, V | L, FCD | METR-LA, TaxiBJ | ✓ | ✗ |
| GSRNN | [91] | Ur | F | FCD | BikeNYC, TaxiBJ | ✓ | ✗ |
| Graph Wavenet | [84] | Fw | S | L | METR-LA, PeMS | ✓ | ✓ |
| 3D-TGCN | [6] | Fw | S | L | PeMS | ✓ | ✗ |
| ST-UNet | [93] | Fw | S | L | METR-LA, PeMS | ✓ | ✗ |
| GaAN | [55] | Fw | S | L | METR-LA | ✓ | ✗ |
| Motif-GCRNN | [88] | Ur | S | FCD | TaxiChengdu | ✗ | ✗ |
| STGi-ResNet | [85] | Ur | F | FCD | Didi Chengdu | ✓ | ✗ |
| T-GCN | [94] | Fw, Ur | S | FCD | SZ-taxi, Los-loop | ✗, ✓ | ✗ |
| FlowConvGRU | [97] | Ur | F | FCD | TaxiNYC, TaxiCD | ✓ | ✗ |

Table 3: Graph Neural Networks approaches to traffic forecasting. For each publication, we list the scope of application (*Ur*: urban, *Fw*: freeway) , the variables predicted (*S*: speed, *F*: flow, *V*: volume), the data source (*L*: loop counters, *FCD*: floating car data), the datasets used for experiments, whether or not these datasets are open, and whether there exists open-source code implementations of the models.

Attention mechanisms have also shown to be a promising approach to spatio-temporal forecasting. In addition to the GaAN architecture [55] described in Section 3.1.3, Attention Based Spatial-Temporal Graph Convolutional Networks (ASTGCN) [8] combine attention mechanisms on both the spatial and the temporal components with a multi-view approach on temporal patterns (namely, recent, daily-periodic and weekly-periodic). Figure 9 illustrates the intuition on how attention mechanisms can help make better predictions as well as output more interpretable models.

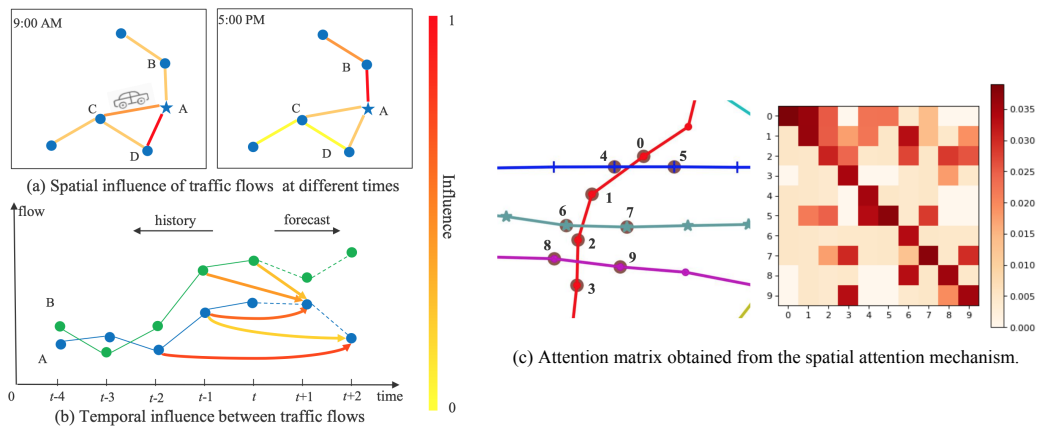

Figure 9: (a) and (b) Spatial-temporal correlation diagram of traffic flow. (c) Attention matrix of sub-graph with 10 detectors, where the *i*-th row indicates how strong the time series of node *i* correlates with every other node. As an example of the interpretability aspect of this model, we can see that it indicates that pairs of nodes most strongly correlated tend to be spatially close, such as 1 and 6, 5 and 4, and 9 and 3. [8] Reprinted with permission.

Another rich approach has been to enable models to learn the graph adjacency matrix from the data and its patterns such as time series similarity between different nodes instead, or in addition to, the usual spatial distance. Variants of this approach are proposed by models that include Dynamic Spatio-Temporal Graph-based CNNs (DST-GCNN) [82], 3D Temporal Graph Convolutional Networks (3D-TGCN) [6], Graph WaveNet [84] and FlowConvGRU [97].

Finally, other models focus on leveraging other approaches such as the sparsity of traffic data [91], pooling and unpooling layers [93] or unsupervised learning of node representations [86].

4.2 Datasets

In spite of the differences described above, nearly every model has been experimentally evaluated in similar datasets with a similar configuration. In particular, the timestep of the datasets is, with few exceptions, 5 minutes, and the prediction horizon is 15, 30 and 60 minutes (that is, 3, 6, and 10 timesteps). The extension of the dataset is also very often around 4 months, which is equivalent to about 35000 timesteps.

Two datasets in particular standout as the most used for benchmarking: METR-LA and PEMS-BAY. These datasets contain traffic information collected from loop sensors at various locations of two networks of freeways in California, one in Los Angeles, the other in the Bay Area. Typically only a small fraction of the data is used. For example, data is aggregated in 5 minutes interval and only between 200 and 1000 sensors are used to test the models.

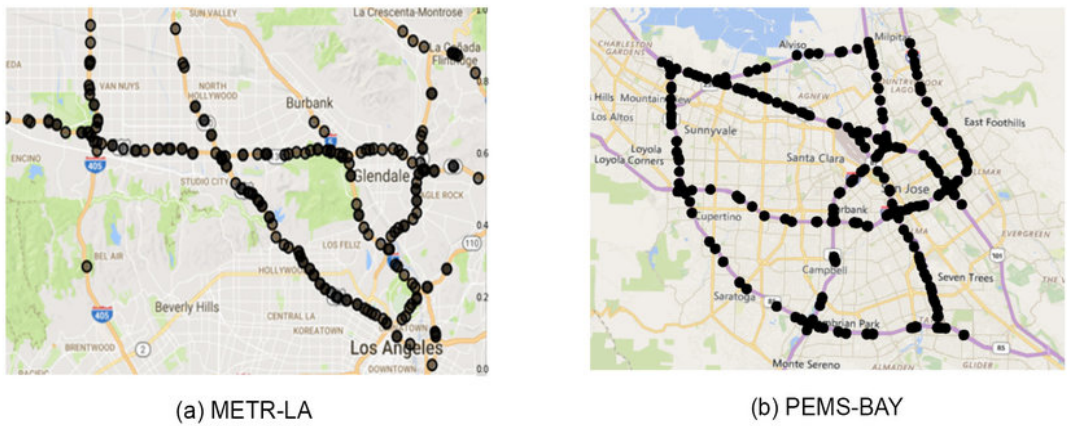

(a) METR-LA                    (b) PEMS-BAY

Figure 6: Sensor distribution of the METR-LA and PEMS-BAY datasets [83]. Reprinted with permission.

## 5 Challenges and research opportunities

While the results obtained in various traffic forecasting tasks by deep learning models and, notably, graph neural networks have been very successful, there are still a number of open issues and traffic forecasting is still an unsolved problem. These challenges constitute a rich set of research and engineering opportunities [3], [98] which include integrating in a systematic way exogenous factors (such as road accidents and weather), designing more sophisticated evaluation metrics, integrating traffic forecasting with other downstream applications, going from volume or speed prediction to travel time prediction, improving the interpretability of the models and moving from prediction to causation. Additionally, ongoing work tackles the problem of developing Bayesian methods that provide adequate predictions beyond point estimates (such as confidence intervals) and tackling the issue of data ageing and concept drift by developing models that take into account when road segments or bus lines are added or removed from the network.